\documentclass{article}

\usepackage{arxiv}
\usepackage[utf8]{inputenc} 
\usepackage[T1]{fontenc}    
\usepackage{hyperref}       

\usepackage{url}            
\usepackage{booktabs}       
\usepackage{amsfonts}       
\usepackage{amsmath}
\usepackage{nicefrac}       
\usepackage{microtype}      

\usepackage{graphicx}
\usepackage{natbib}
\usepackage{doi}

\newcommand\addtag{\addtocounter{equation}{1}\tag{\theequation}}

\usepackage{graphicx}
\usepackage{latexsym}

\usepackage{booktabs}
\usepackage{multirow}
\usepackage{xcolor}

\renewcommand{\th}[1]{}
\newcommand{\argmin}{\texttt{argmin}}

\newcommand{\glsf}[1]{\glsdohypertarget{glo:#1}{\acrshort{#1}}\glsunset{#1}}
\newcommand{\eqa}[1]{\begin{equation}#1\end{equation}}
\newcommand{\eqn}[1]{\begin{equation*}#1\end{equation*}}

\usepackage{float}
\usepackage{subcaption}

\usepackage[acronym]{glossaries}

\usepackage{glossaries-extra}
\setabbreviationstyle[acronym]{long-short}

\glssetcategoryattribute{acronym}{nohyperfirst}{true}

\newacronym{iot}{IoT}{Internet of Things}
\newacronym{ar}{HAR}{Human Activity Recognition}

\usepackage{algorithm}
\usepackage{algpseudocode}

\def\tsc#1{\csdef{#1}{\textsc{\lowercase{#1}}\xspace}}
\tsc{WGM}
\tsc{QE}

\title{Meta-Decomposition: Dynamic Segmentation Approach Selection in IoT-based Activity Recognition}
\usepackage[noabbrev,capitalise]{cleveref}

\author{
Seyed M.R. Modaresi\\
LIPN-UMR-CNRS 7030, \\
Sorbonne University Paris Nord,\\ Paris, France\\
\texttt{modaresi@lipn.univ-paris13.fr} \\
\And
Aomar Osmani \\
LIPN-UMR-CNRS 7030, \\
Sorbonne University Paris Nord,\\ Paris, France\\
\texttt{ao@lipn.univ-paris13.fr} \\
\And
Mohammadreza Razzazi\\
Computer Engineering Department, \\
Amirkabir University of Technology,\\ Tehran, Iran\\
\texttt{razzazi@aut.ac.ir}  \\
\AND
Abdelghani Chibani\\
Laboratory of Images, Signals and Intelligent Systems\\ Université Paris-Est Créteil,\\ Paris, France\\
\texttt{achibani@gmail.com}
}
\begin{document}

\maketitle

\begin{abstract}
    Internet of Things (IoT) devices generate heterogeneous data over time; and  relying solely on individual data points is inadequate for accurate analysis.
    Segmentation is a common preprocessing step in many IoT applications, including IoT-based activity recognition, aiming to address the limitations of individual events and streamline the process. However, this step introduces at least two families of uncontrollable biases. The first is caused by the changes made by the segmentation process on the initial problem space, such as dividing the input data into 60 seconds windows. The second category of biases results from the segmentation process itself, including the fixation of the segmentation method and its parameters.
    To address these biases, we propose to redefine the segmentation problem as a special case of a decomposition problem, including three key components: a \textit{decomposer}, \textit{resolutions}, and a \textit{composer}.
    The inclusion of the composer task in the segmentation process facilitates an assessment of the relationship between the original problem and the problem after the segmentation. Therefore, It leads to an improvement in the evaluation process and, consequently, in the selection of the appropriate segmentation method.
    Then, we formally introduce our novel \textit{meta-decomposition} or \textit{learning-to-decompose} approach. It reduces the segmentation biases by considering the segmentation as a hyperparameter to be optimized by the outer learning problem. Therefore, meta-decomposition improves the overall system performance by dynamically selecting the appropriate segmentation method without including the mentioned biases. Extensive experiments on four real-world datasets demonstrate the effectiveness of our proposal.

\end{abstract}


\section{Introduction}
%
\label{sec:intro}
We are moving towards the \gls{iot}, and the number of deployed sensors is rapidly increasing. These sensors collect uni-modal or multi-modal observation data such as temperature sensor, door sensor, optical sensor \citep{Kumar2023,Perera2014,Messaoud2020}.
Recognition of human activities (\glsf{ar}) from \gls{iot} data is expected to be the heart of myriad \gls{iot} applications such as healthcare, smart homes, and security \citep{Mocrii2018,Qian_Pan_Miao_2021_2,chen2021deep,Perera2014,Kumar2023,Messaoud2020}.
Particularly, with the growing population of older adults which is expected to double by 2050, there has been a surge of interest in sensory and processing data for \gls{ar} to support their healthcare needs \citep{Ariza-Colpas2022,Sokullu2020}.
\gls{ar} plays a crucial role in enabling devices to monitor, analyze, and enhance the daily lives of humans by recognizing their behaviors \citep{chen2021deep}.
However, it is still a challenging task to provide sufficiently robust activity recognition systems in a real environment due to the complexity and diversity of human behaviors that vary from person to person \citep{Najeh2022,Bouchabou2021}.

\gls{iot} devices generate heterogeneous series of data from sensors located in physical spaces or carried by people, such as smartwatches \citep{Hu2017,Ariza-Colpas2022,Hu2019}.
Unlike transactional data, these series of data are characterized by their continuous and relative nature, making them dependent on their previous and subsequent data points  \citep{Fu2011,ActiveLearning.Cook2015};
Segmentation is a common and essential bias that partitions this long, potentially infinite  data sequence into a set of smaller and more meaningful finite segments. It overcomes the limitation of a single sample and provides adequate information about an activity \citep{chen2021deep,Ni2015,Demrozi2021}.
Segmentation helps to deal with the complexity of \gls{ar} problems, although it directly impacts the recognition quality \citep{Ni2015}.
On one hand, inadequate information in one segment may cause the activity to be poorly detected; on the other hand, if a segment contains too much information, extra complexity may be added for future data processing \citep{Krishnan2014,Ni2015}.
Consequently, a trade-off exists between the sufficiency of information in each segment, minimizing the number of segments, and reducing the processing complexity of each one to discover the expected concepts.
Additionally, since segmenting data alters some characteristics of the sensor's data, it should be carefully considered.

In machine learning, the term ``bias'' refers to any factor that favors one generalization over another  \citep{Mitchell1980,Gordon1995}. The segmentation process introduces at least two families of uncontrollable biases.
The first one is introduced to the model due to the changes in problem space by the segmentation.
For instance, a common approach in \gls{ar} consists of segmenting
the data and feeding them to the model to identify the activity in each segment \citep{ActiveLearning.Cook2015}.
It is often assumed that the classifier performance over the segments follows the whole system performance  \citep{Qian_Pan_Miao_2021_2,ActiveLearning.Cook2015,Bouchabou2021,Perera2014,Bilen2020}.
This hard hypothesis may misleadingly present convincing results because different segmentation algorithms generate various kinds of segments with different sizes and structures. Consequently, the results of those algorithms are in distinct spaces and cannot be compared directly. This issue is explained in detail in \cref{sec:Performance}.
Additionally, activities have some time-related properties \citep{Modaresi2022PAKDD}
that may be lost due to segmentation. For example, steady recognition of the sleeping activity is critical; otherwise, it may incorrectly present a disorder \citep{Alemdar2015}.

The segmentation approach itself is the other bias of the segmentation process. It is often implicitly incorporated with the prior knowledge or assumptions originating from the developers, researchers, or experts during the selection and tuning of the segmentation approach.
Accurate recognition of activities (particularly the complex ones) highly depends on the segmentation method \citep{chen2021deep,Ni2015};
for instance, by including a bias to have smaller segments than the required segment size, the machine learning process may not properly identify the activity.
Meta-learning and AutoML techniques has gained attention in recent years for algorithm and hyperparameter selection.
However, meta-learning requires the availability of multiple tasks to learn from \citep{Aguiar2019,Rossi2021,Huisman2021}, while it is not the case in a single \gls{ar} task, making it challenging to apply meta-learning approaches that depend on learning from multiple related tasks.
In addition, AutoML techniques do not change the algorithm dynamically over time \citep{Mu2022,Taylor2018}. IoT systems is Dynamic in Nature \citep{Kumar2023}, therefore,
it is unrealistic to assume that individual activities remain unchanged; for instance, the daily activities in winter are different from those in summer \citep{chen2021deep}.
Therefore, an appropriate segmentation approach in one period may not be efficient for another one.


In this paper, to address these biases in the segmentation preprocessing step in the \gls{ar} system, we first reformulate the segmentation as a decomposition problem and then introduce our novel \textit{meta-decomposition} approach to address these biases.
Therefore, the segmentation problem is redefined as a particular case of data decomposition one that includes the decomposer (traditional segmentation), the resolutions (ML), and the composer steps.
The composer step transforms the ML results to the global problem results to better describe and evaluate the impact of the introduced biases in the segmentation process. It addresses the first family of biases.
\\
To overcome the second family of biases, we propose a novel approach called  \textit{meta-decomposition} or \textit{learning-to-decompose} that learns how to decompose the original task (recognizing activities from long data) into smaller sub-tasks. Therefore, it can be integrated with meta-learning techniques that require multiple tasks to improve recognition performance.
Meta-decomposition seeks to reduce the segmentation biases and optimize the overall system performance by learning how to generate sub-tasks rather than assuming the segmentation method as pre-specified and fixed.
In the proposed model, the segmentation is an ML hyperparameter that is learned adaptively
based on the application and constraints
in the outer learning algorithm to improve the recognition quality of the inner learning process.
As explained before, without considering the meta-composer part, meta-decomposition introduces an additional bias in the comparison of different segmentation approaches due to the inconsistency in the segments.
In the experiments,
we propose a simple and effective data-driven approach to demonstrate the feasibility of finding a proper segmentation method and its hyperparameter in our proposal and show the superiority of our approach compared to the other approaches with their best hyperparameters on four public datasets.

The structure of this paper is as follows: Section 2 introduces the segmentation problem, summarizes existing segmentation approaches, and briefly discusses meta-learning and AutoML. Section 3 provides a depiction of the proposed segmentation formulation and the meta-decomposition model. In Section 4, we present the experimental environment, datasets, and our framework, and demonstrate the effectiveness of the meta-decomposition concept. Finally, Section 5 presents the conclusions drawn from this study.


\section{Background and Related Work}
\label{sec:related works}
In sensor-based \gls{ar}, the  objective is to identify activities including both activity class and their temporal duration based on a sequence of input sensor events \citep{ActiveLearning.Cook2015,Mocrii2018}. The information provided by a single sensor event is inadequate for identifying a particular activity. Accordingly,
it is crucial  to partition sensor events into a collection of segments that can be mapped to a specific activity \citep{MinhDang2020,Bouchabou2021,Demrozi2021}.
Segmentation can significantly affect the system performance because it alters some characteristics of the underlying data.
%
Thus, a trade-off exists between the count of segments, the amount of information conveyed by each segment (e.g., segment size), and the processing complexity entailed in each segment \citep{Ni2015}.

Since segmentation can significantly affect the final results, several studies in \gls{ar} mainly work on utilizing pre-segmented sequences \citep{Najeh2022}. Nevertheless, this approach is not practical in real-world scenarios \citep{Krishnan2014,De-La-Hoz-Franco2018a,Bouchabou2021,Xu2020,Recognizing2018}.
Therefore, some studies rely on a segmentation approach that is based on temporal information \citep{Bouchabou2021,Fu2011,Krishnan2014,Ni2015}, similarity or dissimilarity between segments \citep{Najeh2022,Yala2017,Wan2015,Sfar2018}, ontology and domain knowledge \citep{Triboan2019,Sfar2018,Wang2018,Najeh2022}, learning the segment size \citep{Krishnan2014,Sfar2018}, sensor events \citep{Ni2015,Bouchabou2021,OrtizLaguna2011}, activity or explicit segments \citep{Ni2015,Bouchabou2021,Lunardi2018}, gathering sufficient features \citep{Ni2015,Krishnan2014,ActiveLearning.Cook2015,Yala2017,Sfar2018},  {evolutionary computation} \citep{Tak-ChungFu2001}, detecting change points \citep{Aminikhanghahi2019,Zameni2019}, {feasible space window} \citep{Hu2017}, and hybrid approaches \citep{Najeh2022}. It has been proved that dynamic segmentation approaches perform better than static ones \citep{Fu2011}.
However, the aforementioned studies are designed to be used in a particular application and dataset. e.g., some of them need continuous senses, meaning that all sensor values must be available at each time point; others work on sparse sensor streams, where sensor events are triggered only because of human activities, like motion sensor sequence \citep{Krishnan2014}.

Regarding recent surveys of human activity recognition in smart homes \citep{Bouchabou2021,Ariza-Colpas2022,MinhDang2020,Wang2021}, the most common approaches for data segmentation are Time Windows (TW), Event Windows (EW), and Dynamic Windows (DW).
Kasteren et. al. \citep{Kasteren2011} determined that a time window of 60s in TW provides a high classification performance for binary sensors, and it has been used as a reference in several recent works \citep{Bouchabou2021,Medina-Quero2018,Hamad2021,Hamad2020}. Moreover, for EW that has variable window duration due to the occurrence of events at various times, a window of 20 to 30 events is commonly selected  \citep{Bouchabou2021,Aminikhanghahi2019}.
However, these parameters are completely dependent on a given dataset, and the  significance of selecting an appropriate one is studied in \citep{MinhDang2020} for window size.
Quigley et. al. \citep{Quigley2018} demonstrate that although TW reaches a high accuracy, it fails to properly identify all classes.
On the other hand, DW uses a non-fixed window size and tries to estimate the activity duration based on the sensor events. However, this approach is inefficient for complex activities \citep{Bouchabou2021,Krishnan2014,Quigley2018,Shahi2017}.




There is a rise in deep learning approaches for \gls{ar} \citep{Wang2019,chen2021deep,Bouchabou2021,Bouchabou2021b,Liciotti2020}.
Guan and Plötz \citep{Guan2017} claim that the deep learning method can be insensitive to the window size in \gls{ar}. Yet, it is not the case for \gls{ar} scenarios with sparse data since the mentioned studies are either on the pre-segmented data \citep{Bouchabou2021b} or their segmentation parameters should be tuned to achieve satisfying performance \citep{chen2021deep}.

To automate algorithm and hyperparameter selection, AutoML techniques are used \citep{Mu2022,Taylor2018}. However, they do not adopt the algorithm regarding the incoming data.
Meta-learning, often known as learning to learn, has been successfully used for algorithm selection  \citep{Aguiar2019} and provides automatic and systematic guidance on algorithm selection based on the information gained through a set of algorithms on various tasks \citep{Rossi2021}.
Meta-learning is not a novel concept \citep{Schaul2010}; however,
there has been a recent rise in interest in meta-learning \citep{Grefenstette2019}.
Hendryx et. al. \citep{Hendryx2019a} and Aguiar et. al. \citep{Aguiar2019} use a meta-learning approach across different image tasks to select the proper algorithm for generating the mask for a given image \citep{Sun2021}. The fundamental reason for this study is the fact that previous studies consider each experience as an independent instance and do not consider the decomposition of each \gls{ar} task into sub-tasks. Additionally, in \gls{iot}, the input sensor events in each experience are characterized by their continuous and relative nature; therefore, they are not independent \citep{Fu2011,ActiveLearning.Cook2015}.

To the best of our knowledge, no work has been done on dynamically selecting the appropriate segmentation method and its hyperparameter over time, despite  the possibility that the appropriate approach and its corresponding hyperparameter for one period may not be optimal for another period \citep{Rossi2021,Adam2019}. Furthermore, the previous studies never view the segmentation problem as a meta-decomposition one.
In the next section, we propose our novel approach to resolve these concerns.


\section{Proposed Model}


Our approach consists of two major steps: (a) redefining the segmentation problem as the data decomposition problem and (b) formalizing our novel meta-decomposition approach.
In the following, we first introduce the definition and terminology and then elaborate on each step.

Following the notations of \citep{Schaul2010,Huisman2021,Hospedales2022}, let us consider domain $A$ as a set of experiments.
Experience $s\in A$ is a broad term used on both supervised and non-supervised learning problems that may refer to an input-target tuple, a single data point, a sequence of events (e.g., in \gls{iot}), etc.
We define $m_\theta$ as a task with hyperparameter $\theta$ (includes, e.g., the initial model parameters, choice of the optimizer, and learning rate schedule) that takes the input experiments and outputs the target activities based on the constraints, objectives, etc. $\mathcal{\phi}(m_\theta,s)$ is associated with measuring the performance of task $m_\theta$ on the experiment $s$.
We denote by $\mathcal{L}(m_\theta,S)$ the expected performance of task $m_\theta$ on $S\subseteq A$, such that:

\eqa{
    \mathcal{L}(m_\theta,S) = \mathbb{E}_{s\in S} \left[\mathcal{\phi}(m_{\theta},s) \right]
}

Measurement $\phi$ is as diverse as the application domains. For instance, in supervised learning, $\phi$ might be the differences between task outputs and teacher-given values \th{or the required time to train or obtain a specific behavior }\citep{Schaul2010}.
Without losing generality, we consider the optimum hyperparameter minimizes the expected performance  ($\theta^{*}_{S}=\argmin_{\theta} \mathcal{L}(m_{\theta},S)$).
Finding globally $\theta^*$ is computationally infeasible \citep{Huisman2021}. {Therefore, we approximate it ($\theta^{*}_{S}\approx g_{\omega}(S)$) guided by pre-defined meta-knowledge $\omega$ which includes, e.g., the initial model parameters ($\theta_0$), choice of the optimizer, and learning rate schedule \citep{Hospedales2022}.
}

In sensor-based \gls{ar}, each experiment contains a sequence of various sensors occurrences, and activities information (such as their label and duration); task $m_\theta$ refers to the activity recognition model and its hyperparameters; and $\phi(m_\theta,s)$ evaluates the performance of the activity recognition model on the experiment $s$. For instance, $\phi$ can evaluate the duration of correctly identified activities.

\subsection{Decomposition}

Even though decomposition is a well-known approach in designing algorithms \citep{CLRS2009}, the data segmentation problem has never been viewed as a data decomposition problem, which consists of a decomposer that splits the input sequence into a set of smaller data (traditional segmentation), resolutions that find the concepts from these smaller segments (usually less complex than original resolutions), and a composer that combines the sub-results to generate the overall results.

To the best of our knowledge, previous studies in the literature have not considered the composer component  \citep{Krishnan2014,Qian_Pan_Miao_2021_2,Yala2017,De-La-Hoz-Franco2018a,Cumin2017,Wang2019,chen2021deep,Hussain2019,Bernard2018,Viard2018a,}
These studies have made the implicit assumption that the segmentation process preserves the integrity of the whole problem, and as such, the overall system performance is assessed based on the output of each segment. This hard hypothesis may misleadingly present convenient results without even reducing the problem complexity, which is explained in \cref{sec:Performance}. Therefore, this study explicitly redefines the segmentation problem as a data decomposition that incorporates all three components: the decomposer, the resolutions, and the composer.
The introduced biases, loss of information, and performance of each component can affect the overall performance of the system and should be carefully evaluated.
Furthermore, neglecting the composer component would lead to inconsistencies in the comparison of different segmentation algorithms, as it will be discussed in detail in \cref{sec:Performance}. Therefore, including the composer component in the segmentation problem is crucial for accurately evaluating and comparing the performance of various segmentation algorithms.
%

Accordingly, the decomposer task ($d_\delta$ parameterized by $\delta$) decomposes $m_\theta$ into $M$ resolution sub-tasks ($\Pi_{\Psi}=\{\pi^{i}_{\psi_i}\}_{i=1:M}$ such that each sub-task is parameterized by $\psi_i\!\in\!\Psi$), and the composer task ($c_\sigma$ parameterized by $\sigma$) that combines the results of sub-tasks to produce the overall system results.
Task $m_\theta$ is decomposable under the measurement $\mathcal{L}$  to  $\Pi_{\Psi}$ and $c_{\sigma}$ if and only if the composition of sub-tasks does not
perform worse up to $\epsilon$ than task $m_\theta$. Formally:
\eqn{\label{eq:decomposablity}
d_{\delta,\mathcal{L}}(m_\theta)=\Pi_{\Psi},c_{\sigma}\  \iff\ \addtag  \mathcal{L}(c_{\sigma}(\Pi_{\Psi}),A) - \mathcal{L}(m_{\theta},A) \leq \epsilon
}
The task's performance after decomposition can not surpass the original task due to the severed relationship between events, leading to information loss.
When $\epsilon$ is zero, it means that $m_\theta$ is strongly decomposable to $\Pi_{\Psi}$ and $c_{\sigma}$ without any loss of information, otherwise, it is a weakly decomposable task, and some information is lost. $\delta, \Psi$, and $\sigma$ show the dependencies of the model on pre-defined assumptions about the decomposition, for example, the segmentation approach such as time window and event window, and their internal parameters such as window size. These assumptions can affect the global system's performance.
For example, the optical character recognition (OCR) task is commonly decomposed into the sub-tasks with sub-images (that each contains one character), and the composition task merges the results of those tasks to produce the whole problem result (full text).
\Cref{eq:decomposablity} shows that the decomposability of a task depends on the task target, i.e., on the objective function measured by $\mathcal{L}$.
For instance, while we can decompose the face recognition task to analyze only the color frame task with an accuracy of 99.5\% \citep{Lu2021}; this decomposition may be inadequate in a highly secure application, where a more detailed decomposition involving depth and color sub-tasks may be required \citep{Apple2022}.

Obviously, all the mentioned components (decomposer, resolution, and composer) play a crucial role in measuring the performance, the introduced biases, loss of information, and complexity of the whole system and should be considered in designing and evaluating segmentation approaches. In particular, ignoring the composer component leads to inconsistencies and difficulties in comparing different segmentation algorithms, which are elaborated more in \cref{sec:Performance}.

\subsection{Meta Decomposition}
\begin{figure*}[t]
    \centering
    \includegraphics[width=.9\textwidth]{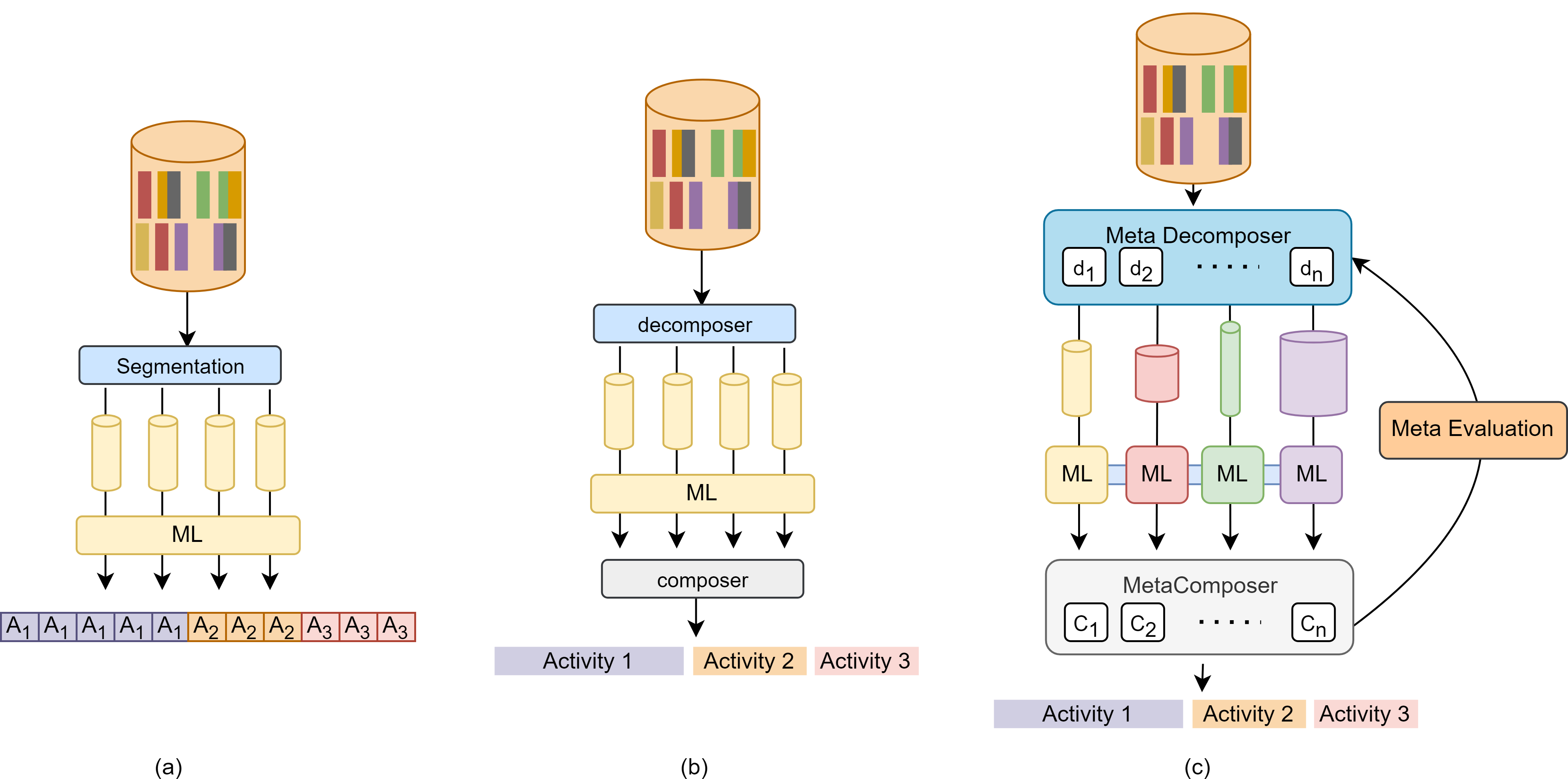}
    \caption{(a) shows the conventional segmentation approach, which creates a set of segments in data preparation and then treats them as individual instances to be inputted into the ML system (abstract schema from \citep{Kumar2023}). (b) illustrates the proposed formulation of segmentation as a data decomposition problem, including the decomposer, the ML model, and the composer. (c) provides an overview of our proposed meta-decomposition model. It can dynamically select the proper decomposition sub tasks from multiple decomposition algorithms.}
    \label{fig:describe meta}
\end{figure*}

In a traditional segmentation, the probability distribution of data is supposed to be unknown but stationary. Nevertheless, the underlying distribution of data in real-world \gls{iot} systems naturally changes over time \citep{Rossi2021}. Additionally, fixing the segmentation and its hyper-parameter is the second family of biases (\cref{sec:intro}) that often implicitly incorporated with the prior knowledge or assumptions
originating from the developers, researchers, or experts.
\\
Therefore, we propose our novel meta decomposition approach to resolve these concerns. Learning-to-decompose or meta-decomposition is defined as a model that can dynamically and systematically select and tune the decomposition algorithm.
Formally, in \cref{eq:meta}, we consider  $d^{i}_{\delta_i}\in \mathcal{D}$ as the i-th decomposer task which generates sub-tasks $\Pi^{i}_{\Psi_{i}}$ and the composer task $c^{i}_{\sigma_{i}}$; then, we define the meta-decomposition task ($\widehat{d}_{\hat{\delta}}$) as  the selection of sub-tasks ($\widehat{\Pi}_{\hat{\Psi}}$) and the meta-composer task ($\widehat{c}_{\hat{\sigma}}$), such that in the meta-evaluation, the meta-decomposition task outperforms those decomposition tasks individually.
\eqn{
\label{eq:meta}
\widehat{d}_{\hat{\delta},\mathcal{L}}(m_
    {\theta})=\widehat{\Pi}_{\hat{\Psi}}, \widehat{c}_{\hat{\sigma}},\addtag\\
\quad
s.t., \widehat{\Pi}_{\hat{\Psi}}\subseteq \underset{i}{\cup}\ \Pi^i_{\Psi_i} \land
\mathcal{L}(\widehat{c}_{\hat{\sigma}}(\widehat{\Pi}_{\hat{\Psi}}),A)\!-\!\underset{i}{\min\ } \mathcal{L}(c^i_{\sigma_i}(\Pi^i_{\Psi_i}),A)\!<\!0
}
This definition is illustrated in \cref{fig:describe meta}. The meta-decomposition task can be carried out in several ways to efficiently and dynamically select the proper segmentation approach and its hyperparameters depending on the incoming data, application and constraints.
Moreover, the proposed model can be easily extended to the arbitrary number of meta-levels and is not limited to a single layer of meta-decomposition. For example, meta-meta-decomposition algorithm can generate the sub-tasks for the inner meta-decomposition algorithm.

As it is mentioned before, meta-learning \citep{Schaul2010,Aguiar2019,Vanschoren2019} has been successfully used for algorithm selection over various tasks (e.g., \citep{Sun2021}); however, it required multiple tasks and is not applicable for one single \gls{ar} task.
In meta-decomposition, we generate the sub-tasks from a single task. These sub-tasks can be fed to the meta-learning approaches \citep{Schaul2010}. Therefore, we can enhance the overall system performance without including hard prior biases about the fixation of the segmentation method and its hyperparameters. The next section describes more about it with an experiment.

The current definition is open to interpretation regarding the distinction between decomposition and meta-decomposition. Specifically, combining a decomposition with a meta-decomposition, as well as any number of further meta-meta-decompositions, can always be seen as a single "flat" decomposition algorithm.
On the other hand, some decomposition methods can be seen as a type of meta-decomposition. For instance, the one that decomposes data and changes the window size based on the input data while composing the results can be considered a basic form of meta-decomposition.


\section{Experiment}
In this paper, we proposed the meta-decomposition to improve the overall performance of the system where the change in the environment is inevitable and a decomposer task in one period may become inappropriate in another period. In other words, meta-decomposition adaptively learns the proper decomposition for different data in dynamic environments \citep{Rossi2021}.

In the experiments, we demonstrate the meta-decomposition effectiveness in \gls{iot} data in \gls{ar}. To select appropriate sub-tasks from the available decomposer tasks, we first decompose this long data into a set of meta-segments; then extract the meta-features of these meta-segments to learn how to select a suitable base decomposition task. We can also extend this step to include a higher-level meta-meta-decomposition task. However, we only consider one level in these experiments for simplicity. The details of these experiments are:
In the following subsections, we detail our experiments, selected datasets, environment, framework, baselines, and evaluation method. Following that, we present a discussion on the results.

\subsection{Experimental Setup}
\begin{figure}[tbh!]
    \centering
    \begin{subfigure}[b]{0.31\textwidth}
        \centering
        \includegraphics[width=\textwidth]{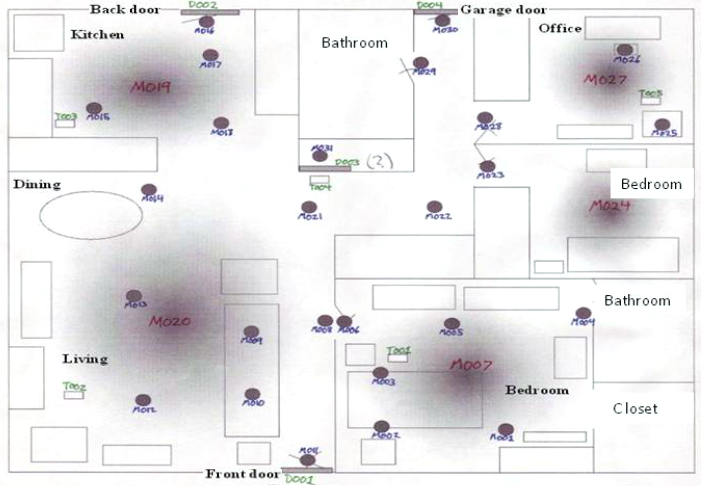}
        \caption[CASAS Aruba dataset sensor configuration]{Aruba. It contains the events from 34 binary sensors for 7 months.}
        \label{fig:5:aruba}
    \end{subfigure}
    \hfill
    \begin{subfigure}[b]{0.31\textwidth}
        \centering
        \includegraphics[width=.9\textwidth]{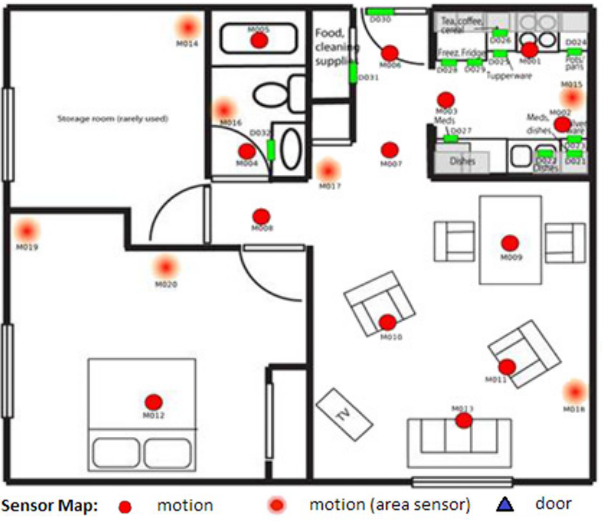}

        \caption{Home1. It contains the events from 32 binary sensors for 5 months.}
        \label{fig:5:Home1}
    \end{subfigure}
    \hfill
    \begin{subfigure}[b]{0.31\textwidth}
        \centering
        \includegraphics[width=.9\textwidth]{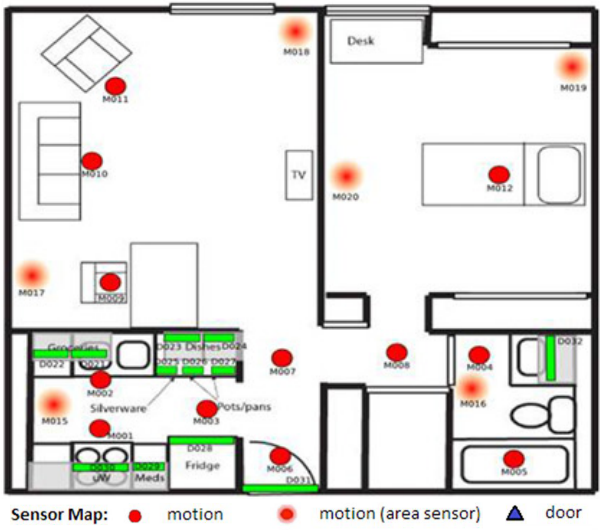}

        \caption{Home2. It contains the events from 30 binary sensors for a period of 5 months.}
        \label{fig:5:Home2}
    \end{subfigure}
    \caption[CASAS Aruba, Home1 and Home2 datasets sensors configuration]{CASAS Aruba \citep{Cook2012}, Home1 and Home2 datasets \citep{Krishnan2014} sensors configuration.  Around 70\% of the activities are unlabeled.}
    \label{fig:5:Comparison}
\end{figure}

\begin{figure}[tbh!]
    \centering
    \includegraphics[width=0.8\textwidth]{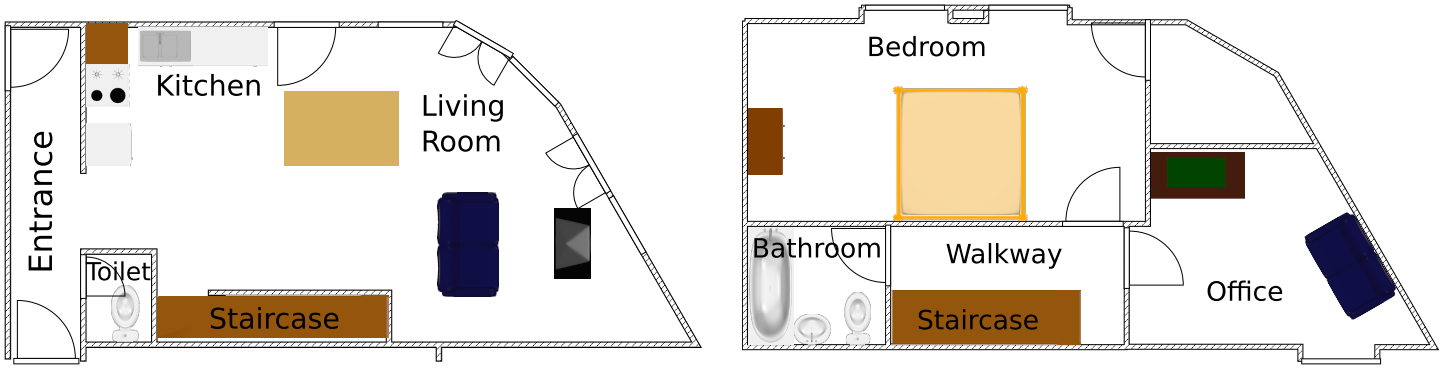}
    \caption[Orange4Home dataset sensor configuration]{Orange4Home Dataset Configuration - This dataset represents a two-floor home with 236 sensors, including 83 binary, 55 integer, 67 float, and 31 categorical sensors. These sensors capture data related to motion, switches, humidity, water consumption, luminosity, temperature, weather conditions, and heater settings over a 5-month period.}
    \label{fig:5:orange4home}
\end{figure}

\begin{figure}[tbh!]
    \centering
    \includegraphics[width=\textwidth]{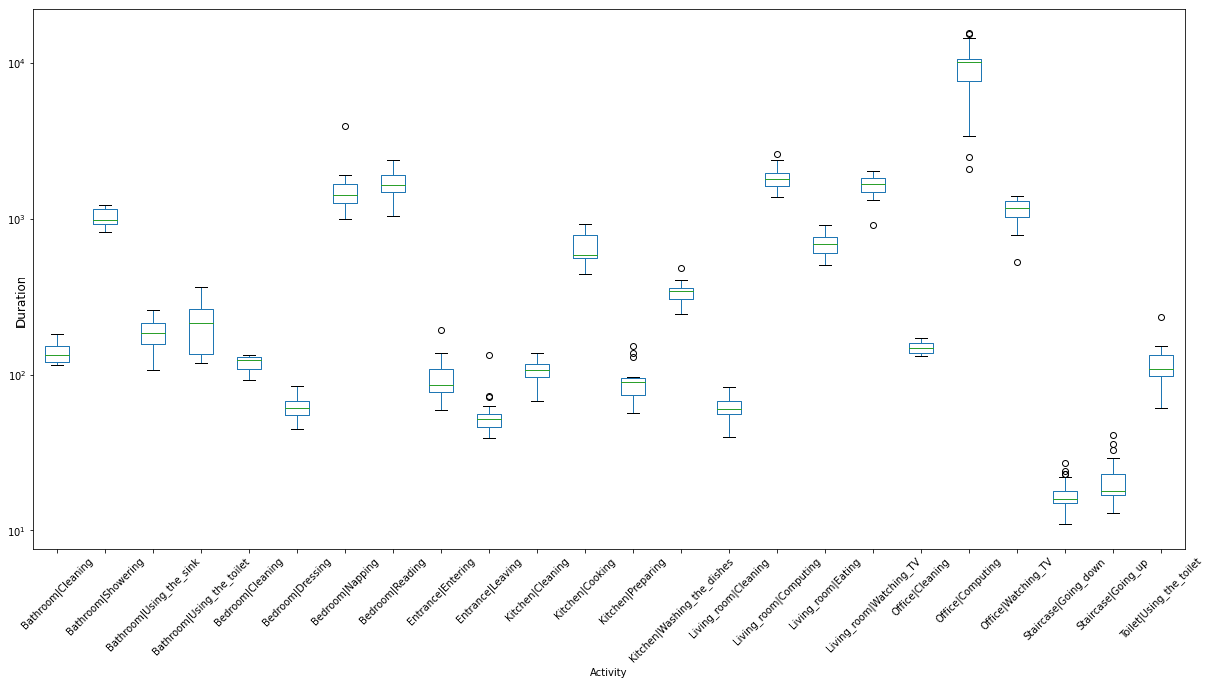}
    \caption[Orange4Home Activity Duration]{Orange4Home Activity Durations: This box-plot depicts the duration of various activities within the Orange4Home dataset. The y-axis measures time in seconds on a logarithmic scale, showcasing the diversity in average activity durations. Activities are listed on the x-axis, with durations ranging from 10 seconds (e.g., 'going down' activity) to approximately 3 hours (e.g., 'computing' activity).}
    \label{fig:Orange4Home activity duration}
\end{figure}

\begin{figure}[tbh!]
    \centering
    \includegraphics[width=\textwidth]{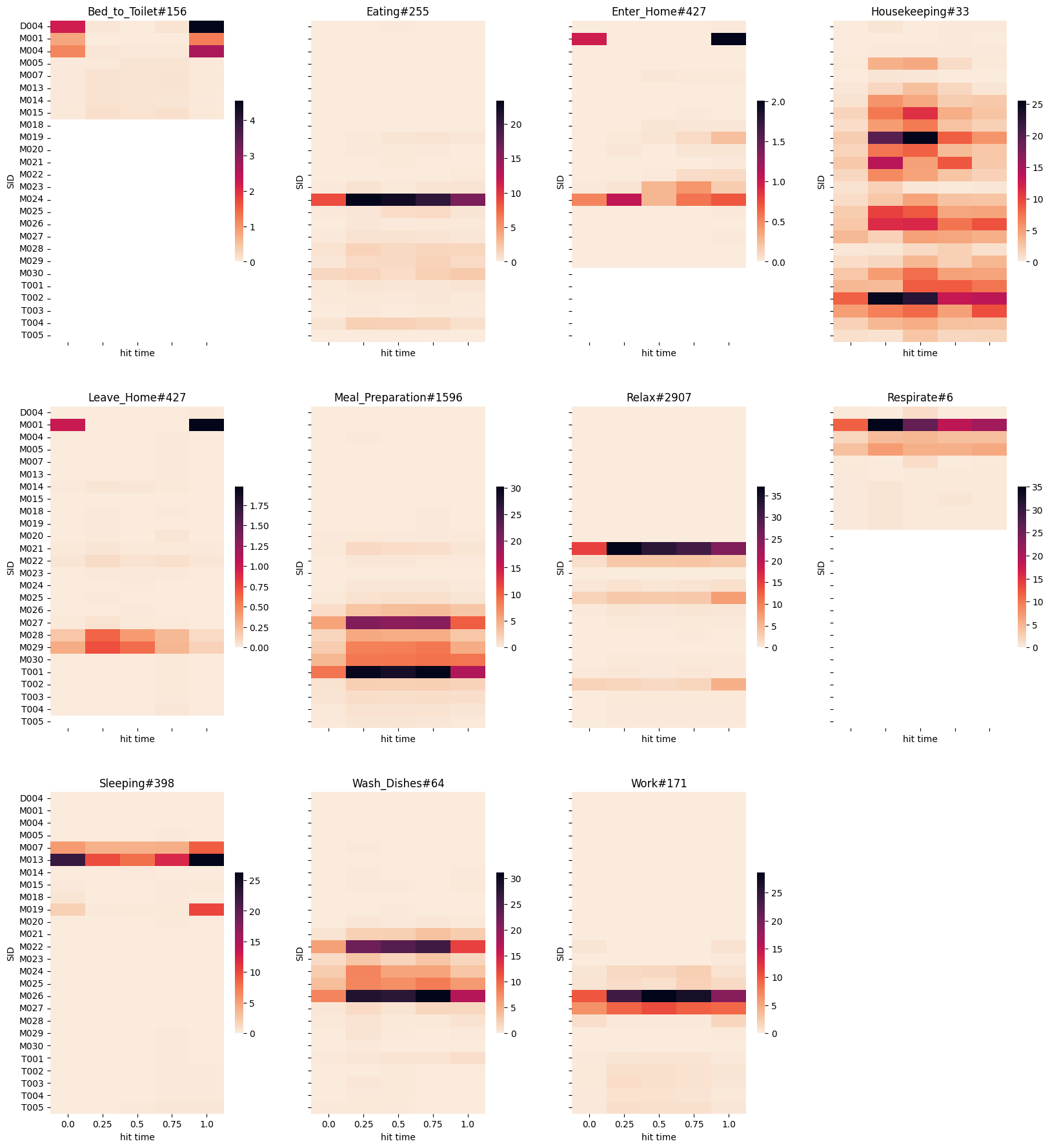}
    \caption[Aruba dataset Sensor Event Frequency]{
        Sensor Event Frequency for the activities in the Aruba Dataset.  For each activity, a hit map is shown which represents the frequency and distribution of sensor events in the Y-axis. The absence of events is shown in white, while darker tones indicate higher event frequencies. On the legend, the number of sensor's heats are shown.
        These visual patterns reveal significant differences in the occurrence and timing of sensor-triggered activities, such as between 'Meal Preparation' and 'Wash Dishes'.
        Pre-segmented data showcases distinct patterns, reducing ambiguity and enhancing the recognition performance of activities. Although valuable for experimental analysis, the practicality of pre-segmented data in real-world scenarios may be limited. Similar observation exists in the other datasets that are represented in the appendix.
    }
    \label{fig:Aruba hit time}
\end{figure}

\begin{figure}[tbh!]
    \centering
    \includegraphics[width=\textwidth]{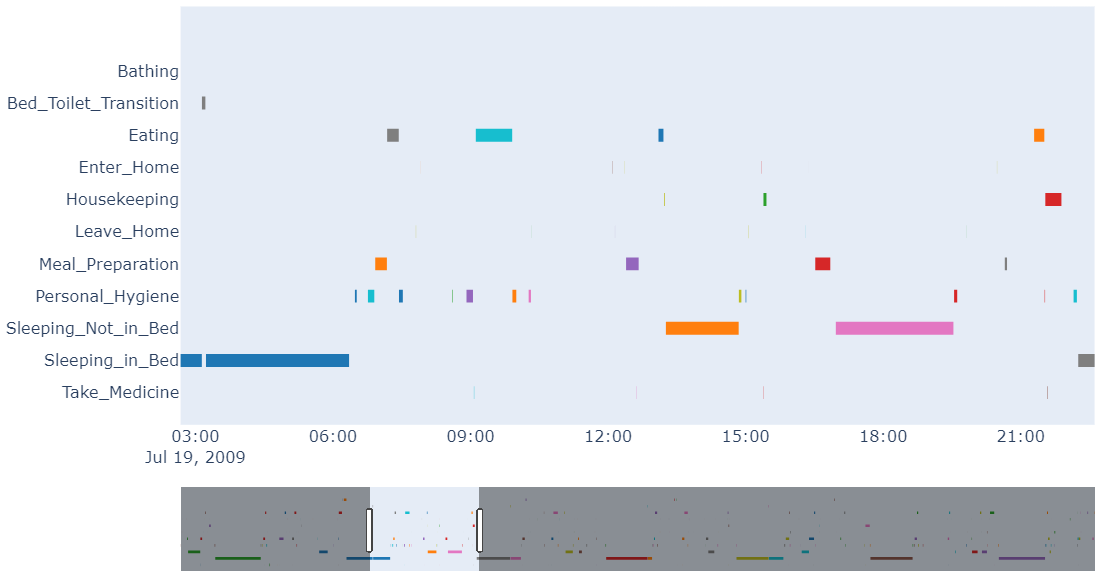}
    \caption[An Illustrative Day from the Home1 Dataset]{Example of Daily Activities from the Home1 Dataset. The pattern of one week is represented by a compact bar at the image's bottom, with the selected day accentuated for emphasis. The y-axis enumerates the different classes of activities, while the horizontal lines mark the length of each activity's occurrence. Notably, the consistency in activity patterns across various days calls for careful analysis to avoid overfitting by overlooking temporal factors. The patterns of other datasets are available in the appendix.}

    \label{fig:home1 activities}
\end{figure}

\begin{figure}[tbh!]
    \centering
    \includegraphics[width=0.6\textwidth]{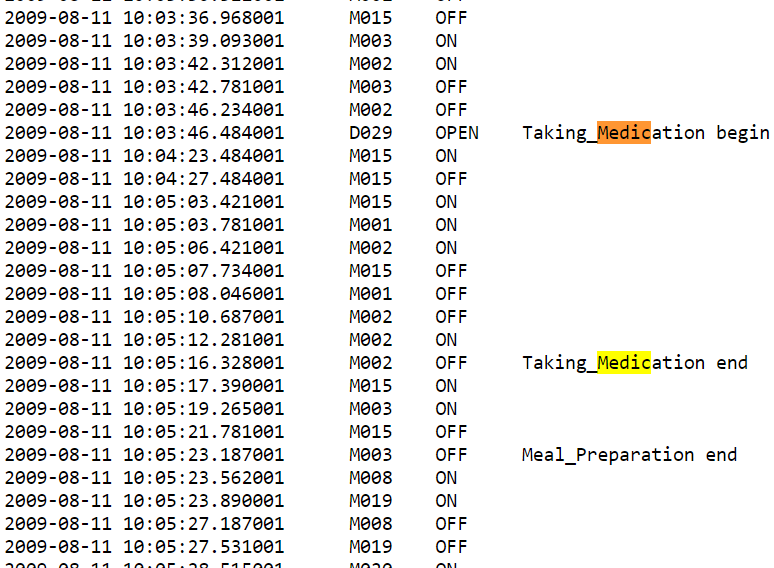}
    \caption{An extract of Home2 dataset with raw data. Within this dataset, annotated sensor events encompass various predefined categories of daily life activities, and the untagged sensor events are consistently labeled as "Other Activity". The start and end terms indicate the initiation and conclusion of activities. In the implementation, categorical information is encoded using a one-hot encoding approach. Within CASAS datasets, events are derived from various types of sensors. Specifically, motion sensors are denoted by IDs starting with "M," door closure sensors by IDs starting with "D," and temperature sensors by IDs beginning with "T."}

    \label{fig:home2 sample}
\end{figure}

\paragraph{\textbf{Datasets:}}
Experiments are conducted  on various  public testbeds, including the widely-used \citep{De-La-Hoz-Franco2018a,Bouchabou2021,Ariza-Colpas2022} WSU CASAS Home1, Home2 \citep{Krishnan2014},
and Aruba \citep{Cook2012}
datasets that have around 32 sensors and between 250,000 to 1,700,000 events, Orange4Home (Orange4H) dataset \citep{Cumin2017} that has 207 sensors and about 700,000 events.
Each testbed consists of heterogeneous sensor events and the daily activities of an individual in a smart apartment.
They have imbalanced activity classes, activity durations, and sensor events. e.g., bathroom activities are frequent and last a few minutes with a few sensor events, while cooking activities may occur once a day, last about an hour, and involve numerous sensor events \citep{Medina-Quero2018}. To enhance comprehension of these datasets, we have visualized them in \cref{fig:5:Comparison,fig:5:orange4home,fig:Orange4Home activity duration,fig:Aruba hit time,fig:home1 activities,fig:home2 sample}. Detailed information about each figure is provided in its respective caption.

\paragraph{\textbf{Environment:}}
All experiments are run on an NVIDIA DGX-1 machine featuring a Tesla V100-32 GPU, Intel Xeon E5-2698v4 CPUs, and 512 GB of RAM. However, our framework works also on a personal computer.

%
%
%
%
%
%

\subsection{Framework Description and Baseline}
\label{sec:framework}
Our pipeline infers activities and their duration from raw sensor data. The pipeline is composed of several stages: data pre-processing, meta-decomposition, feature extraction, classification, and meta-composition.
For each stage, various techniques are implemented in our repository. However, without losing generality in this experiment, we have fixed the parameters of the inner learner to focus on the meta-decomposition.
For the inner learner, a fully convolutional network (FCN) is selected. It outperforms the long-short-term memory (LSTM) networks, while it is significantly quicker \citep{Bouchabou2021,Bouchabou2021b}. It treats sensor events as words and activity sequences as text sentences.{ Therefore, they encode each sensor event as a word containing the sensor name and its value. For example, if a sensor with id 'door1' fire an 'open' event, it will be encoded as  "door1open". Then, based on the frequency of each word, it will be indexed from 1 (index zero is reserved for padding). Then each sequence of sensor events in a window is mapped to an activity. }
A sequential model with three layers of conv1D, batch normalization, and relu activation with 128, 256, and 128 filters, 1D global average pooling, and softmax layers is used. The hyperparameters of the model in the training phase include the batch size of 1024, epochs limit of 100, validation split of 0.2, Adam optimizer, and categorical cross-entropy loss.
%
Afterward, the composition step converts the ML results to the problem space. As explained before, this step is ignored by several studies. This step itself is a challenging problem and directly impacts the result. We demonstrate its importance using a basic combiner that combines overlapped and neighbor windows.

This paper's idea is demonstrated through experiments using a straightforward yet effective method called SWMeta. Although there is potential for multiple higher-level meta-meta-decompositions, this study only focuses on one layer of meta-decomposition, breaking down the data into one-day meta-segments.
Then, we select randomly $J$ meta segments (in this experiment, $J=8$) and use grid search to find the inner decomposer's hyperparameters for each meta-segment. The inner model parameters is then updated with new decomposer. This process is repeated 100 times in this experiment.
Next, for each meta-segment, proper decomposer parameters will be selected by starting from the global knowledge obtained from the previous step to update the local knowledge that is proper for this meta-segment. We add the meta-features from this meta-segment and the selected decomposer parameters to the new train set. For this, we extract the meta-features containing the number of events triggered by each sensor (normalized by the mean and scaling to unit variance) and the spline transformed day of week and month \citep{Eilers1996}.
After training the new model using this new training set, we estimate the proper decomposer parameters for each meta-segments in the test set. Next, we generate the segments, predict the activity of each segment and compose the predicted activities. After that, we apply the meta-composer to generate global problem solutions.
Based on the recent surveys of \gls{ar} in smart homes, TW, EW, and DW (probabilistic \citep{Krishnan2014}) are the most used segmentation approaches \citep{Bouchabou2021,Ariza-Colpas2022,MinhDang2020,Wang2021}.
Therefore, our meta-decomposer selects the decomposer's hyperparameters (segmentation algorithm and its parameters) dynamically among them.
Finally, we use a multi-layer perceptron model with four hidden layers (three sequential dense layers with 16 Relu activations and batch normalizations and one layer with softmax and linear activation) to train our model to estimate the inner segmentation hyperparameters.
The general idea of this algorithm that is inspired from MAML \citep{Finn2017} is shown in the \cref{alg:metadecoposition}.
\renewcommand{\algorithmicrequire}{\textbf{Input:}}
\renewcommand{\algorithmicensure}{\textbf{Output:}}
\begin{algorithm}[t!]
    \caption{Simple Meta-Decomposition (SWMeta) } \label{alg:metadecoposition}
    \begin{algorithmic}[]
        \Require Training dataset $A^{train}$, Testing dataset $A^{test}$
        \Require Hyperparameters $\widehat{\delta}$
        \Comment{e.g., meta-segment size, $\gamma$}
        \Ensure Predicted activities for $A^{test}$

        \State Initialize primary model $M$ and segment decomposer $D$ using $\widehat{\delta}$.
        \State Generate meta-segments for training: $\mathcal{T}=\{\mathcal{T}_1,...,\mathcal{T}_n\}$ from $A^{train}$.

        \While{termination criterion not met}
        \State Sample a batch $B$ of $J$ tasks from $\mathcal{T}$.
        \For{each  $(Z^{train},Z^{val}) = \mathcal{T}_j$ in $B$}
        \State Optimize $D$ for best segmentation of $M$ using $Z^{train}$.
        \State Decompose validation data: $S = D(Z^{val})$.
        \State Update and train $M$ using segmented data $S$.
        \EndFor
        \EndWhile

        \State Initialize meta-feature matrix $X$ and decomposer vector $y$.
        \For{each task $\mathcal{T}_i$ in $\mathcal{T}$}
        \State Optimize $D'$ for best segmentation starting from $D$ on $\mathcal{T}_i$.
        \State Extract meta-features: $F = \text{MetaFeatures}(\mathcal{T}_i)$.
        \State $X$.append($F$), $y$.append($D'$).
        \EndFor
        \State Train model $N$ on the  $(X,y)$.

        \State Generate meta-segments for testing: $\mathcal{T'}$ from $A^{test}$.
        \State Initialize predicted activities list $C$.

        \For{each task $\mathcal{T}_i$ in $\mathcal{T'}$}
        \State Extract meta-features: $F = \text{MetaFeatures}(\mathcal{T}_i)$.
        \State Predict decomposer: $D' = N(F)$.
        \State Decompose task: $S = D'(\mathcal{T}_i)$.
        \State Initialize result list $R$.
        \For{each segment $s$ in $S$}
        \State Predict activity $k$ for segment $s$ using $M$.
        \State $R$.append($k$).
        \EndFor
        \State $C$.append(compose($R$)).
        \EndFor

        \State \Return metaCompose($C$).

    \end{algorithmic}
\end{algorithm}

\subsection{Performance Measurements}
\label{sec:Performance}
Evaluating the model quality is essential to compare and optimize different approaches.
As described in the decomposition definition, the processing performance depends on the decomposer, composer, the number of segments and size (structure), and resolution.
Decomposers generate segments with various sizes and structures. Thus, it is impossible to compare their quality without transforming the results into a unified space.
\begin{figure}
    \centerline{\includegraphics[width=0.7\columnwidth]{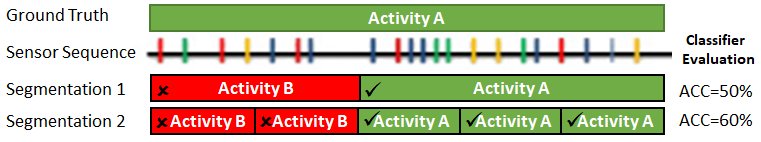}}
    \caption{Comparison of two segmentation algorithms. One of the segments in the first method and two of those in the second method fail to detect Activity $A$ accurately. The box shows the activity and its duration. The vertical colored lines in the sensor sequence represent the activation of various sensors at different time intervals (represented by the horizontal line).
        Correct predictions are denoted by '$\checkmark$' while incorrect ones are denoted by '$\times$'.}
    \label{fig:sample_segmentation}
\end{figure}
\Cref{fig:sample_segmentation} illustrates two examples of \gls{ar} systems that use different segmentation algorithms.
Activity $A$ is not appropriately detected in half of the segments in the first segmentation method, while it is not detected in the 40 percent for the second segmentation method.
Classifier metrics are frequently used to analyze the performance of \gls{ar} systems \citep{Kumar2023,Krishnan2014,Ni2015,ActiveLearning.Cook2015,Fu2011,Qian_Pan_Miao_2021_2,Yala2017,De-La-Hoz-Franco2018a,Cumin2017,Bouchabou2021,chen2021deep,Bernard2018,Viard2018a} However, it may lead to biased results when comparing different segmentation approaches.
For instance, in \cref{fig:sample_segmentation}, the class accuracy is 50\% in the first segmentation method, while it is 60\% in the second one. Obviously, their performances are similar in terms of duration. However, the aforementioned metric fails to represent the situation correctly as the various segmentation approaches can  alter the problem space substantially.
Moreover, activities have some properties
related to their duration \citep{Modaresi2022PAKDD}. For example,
steady recognition of the sleeping activity is critical; otherwise, it may misleadingly present a disorder \citep{Alemdar2015}. However, the segmentation process may break these properties.

\begin{figure}
    \centering
    \includegraphics[width=0.4\columnwidth]{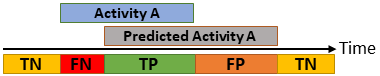}
    \caption{An example TS-CM calculation after composing the classifier results. \tiny{(abbr. T=True, F=False, P=Positive, N=Negative)} }
    \label{fig:TS-CM}

\end{figure}

Therefore, after applying the composition step,
we adopt the time slice (TS) based confusion matrix (CM) \citep{Kasteren2011} to evaluate different segmentation methods in a unified space.
This TS-CM helps us to compare f-score, accuracy, recall (TPR), and other CM measures in an identical space.
\Cref{fig:TS-CM} shows the calculation of TS-CM on an activity.
To obtain the generalized performance, five-fold cross-validation is used for model evaluation, which is a wide approach used for model evaluation in \gls{ar} \citep{Bouchabou2021}.
It splits the dataset into five parts based on temporal occurrence. At each step, four parts are selected for training and the remaining part for testing. Then, we iterate on the parts until all the parts are used for testing.
To preserve the continuous nature of events, the events of each day appear on only one part. Each configuration is repeated five times, and the average and standard deviation of its results are presented.

\subsection{Results and Discussion}
\begin{figure}[t]
    \centering
    \includegraphics[width=\columnwidth]{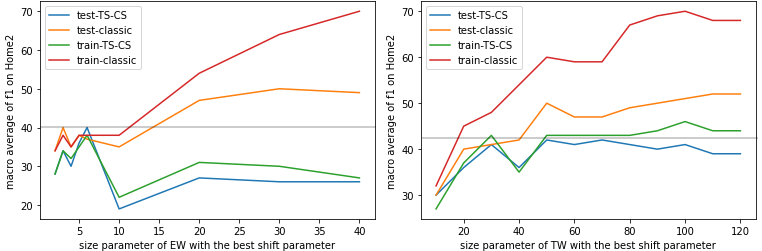}
    \caption{The impact of two segmentation methods (EW and TW) and their parameters (window size and the best shift parameter with that window size) on both train and a test set of the Home2 dataset with classifier metric and TS-CM metric. Interestingly, we can observe that increasing the window size increases the performance measured by the classifier metric, while when we apply the composer to the results and calculate the TS-CM measure, the performance decreases. It demonstrates once more how crucial it is to include the composition component in the segmentation process.}
    \label{fig:home2_size_f1_EW_TW}
\end{figure}

\begin{table}
    \centering
    \caption{Performance evaluation of methods used for segmentation. For the first three methods, the best hyperparameter is selected and shown in parentheses. Our proposed method (SWMeta) dynamically selects the appropriate segmentation method and its hyperparameter at each time. The preliminary results demonstrate that SWMeta outperforms  the other methods alone. These experiments show uncontrollable biases introduced by the segmentation process can be reduced, as is expected in the formulation of meta-decomposition.
        In this table, $w$ and $s$ refer to window size and shift. }
    \label{tab:result}
    \footnotesize
    \begin{tabular}{@{}p{0.3\columnwidth}p{0.3\columnwidth}p{0.2\columnwidth}l@{}}
        \toprule
        Dataset & Segmentor       & TPR       & F1                 \\ \midrule
        \multirow{4}{*}{\shortstack[l]{CASAS Home1                 \\(11 classes)}}    & EW (w=5, s=2)   & 0.65±0.04 & 0.42±0.03          \\
                & TW (w=30, s=20) & 0.48±0.09 & 0.41±0.05          \\
                & DW              & 0.35±0.01 & 0.27±0.01          \\
                & \textbf{SWMeta} & 0.65±0.03 & \textbf{0.43}±0.03 \\\hline
        \multirow{4}{*}{\shortstack[l]{CASAS Home2                 \\(11 classes)}}    & EW (w=6,s=3)    & 0.56±0.08 & 0.40±0.02          \\
                & TW (w=50, s=40) & 0.52±0.07 & \textbf{0.42}±0.04 \\& DW  & 0.32±0.01 & 0.21±0.01\\
                & \textbf{SWMeta} & 0.50±0.12 & 0.39±0.09          \\\hline
        \multirow{4}{*}{\shortstack[l]{CASAS Aruba                 \\(11 classes)}}    & EW (w=3, s=3)   & 0.59±0.05 & 0.34±0.04          \\
                & TW (w=60, s=50) & 0.47±0.06 & 0.33±0.05          \\& DW  & 0.26±0.03 & 0.21±0.01\\
                & \textbf{SWMeta} & 0.61±0.04 & \textbf{0.39}±0.05 \\\hline

        \multirow{4}{*}{\shortstack[l]{Orange4H                    \\(24 classes)}} & EW (w=40, s=20) & 0.27±0.07 & 0.32±0.05          \\
                & TW (w=60, s=60) & 0.32±0.08 &
        0.35±0.06                                                  \\
                & DW              & 0.30±0.01 & 0.34±0.01          \\
                & \textbf{SWMeta} & 0.34±0.04 & \textbf{0.36}±0.03 \\
        \bottomrule
    \end{tabular}
\end{table}

The proposed method is a meta-decomposition method (traditionally segmentation) to be used by recognition methods, not a recognition method in itself.
The impact of the segmentation methods and their parameters are shown in \cref{fig:home2_size_f1_EW_TW}. It highlights the importance of the composition step in the segmentation process and shows that our segmentation reformulation as a decomposition problem improves evaluating of the biases introduced by the segmentation step.
For instance, in the initial subfigure of \cref{fig:home2_size_f1_EW_TW}, it is noticeable that augmenting the size parameter in EW results in an enhancement in performance in the absence of composer (classic), however, it leads to a reduction in performance after the inclusion of the composer (TS-CS). Furthermore, it is worth mentioning that employing 30 events size, as utilized in various researches such as \citep{Bouchabou2021,Aminikhanghahi2019}, introduces a substantial bias. The same applies to TW, thereby necessitating caution in interpreting the results.
To show the usefulness of the meta-decomposition concept, in these experiments, our method learns the appropriate segmentation method each time in the training phase. Then, in the test phase, it selects the appropriate segmentation method and its hyperparameter dynamically at each time. To demonstrate the superiority of this approach, we compare it with the \textbf{best} hyperparameter of each method individually, which heavily
rely on human experience or domain knowledge. The results are summarized in \cref{tab:result}. To find this best parameter for the baseline, we conducted a grid search on each dataset. As we can observe from the table that the recommended window size of 60 seconds for TW in \citep{Bouchabou2021,Medina-Quero2018,Hamad2021,Hamad2020} introduces a bias, as the optimal hyperparameter varies between 30 and 60 seconds for different datasets.
Our proposed approach dynamically selects the best segmentation method at each time among those methods and outperforms those methods individually, except for the Home 2 dataset, which contains few sensor events, thus meta-segment does not have enough data to predict the proper segmentation method and its hyperparameters.

For our inner learning model, we adopt the deep learning model proposed by Bouchabou et. al. \citep{Bouchabou2021b} which is described in \cref{sec:framework}. Assuming identical settings, our results would have been equivalent. However, we introduced three distinct differences in this experiment. First, we include the composition step, which means we rebuild the initial problem results and evaluate the results on that space instead of considering the classification performance, which is more described in \cref{sec:Performance}. As shown in \cref{fig:home2_size_f1_EW_TW}, it produces noticeable disparities.
Second, they assumed that the input sensor events are pre-segmented based on the activity duration; then, they applied the windowing approach to each segment before the deep-learning step, while we do not have such an assumption, which results in the inclusion of significant noise in the learning model. Third,\th{ they generate the segments based on the event window with a shift of one event; then, they use a stratified shuffle on the windows to create the train and test sets. Therefore, some parts of test instances may be included in the training instances. However, before applying the segmentation, we split the dataset to train and test. Four,} we use the macro average while they use the weighted average, which gives more weight to the dominant activities.

These experiments show that segmentation may introduce uncontrollable biases and reduce recognition quality, while, our meta-decomposition concept lessens uncontrollable biases in segmentation by dynamically choosing the proper decomposer and its parameters based on meta-features. They also show that ML can better decompose (traditionally segment) the recognizing activities from sensor data into multiple subtasks without implicitly including knowledge about the problem domain.



\section{Conclusion and Future work}

Segmentation is often considered without the composer part; however, the composer part is an essential step
to reduce the implicit bias 
of the segmentation. Therefore, we redefine the segmentation problem as a data decomposition problem, including a decomposer, resolutions, and a composer.
In addition, while the majority of work in the literature focuses on fixed segmentation approaches that heavily rely on human experience or domain knowledge,
we propose the idea of meta-decomposition or learning how-to-decompose to consider the segmentation process as a hyperparameter inside the outer optimization loop to adaptively select the appropriate one based on the incoming data\th{ by using, for example, a machine learning approach}. It controls and consequently reduces additional biases introduced by this step.
This framework is the first step towards enhancing the \gls{ar} quality without implicitly importing human biases about the application and the used dataset in the algorithms, implementations, and evaluations.
Since the segmentation step changes the problem space and different segmentation algorithms generate heterogeneous segments, this paper discussed and proposed to evaluate them in a unified space.
The experiments demonstrate that the proposed meta-decomposition concept improves the overall machine learning performance and needs to be considered in future studies.
We hope that this work will open the way for proposing more effective meta-decomposition approaches incorporated with meta-learning approaches in our future study.
%

\section*{Declaration of Competing Interest}
The authors declare that they have no known competing financial interests or personal relationships that could have
appeared to influence the work reported in this paper.

\bibliographystyle{unsrtnat}
\bibliography{references}

\appendix
\section{More details on the Used Datasets}

\begin{figure}[tbh!]
    \centering
    \includegraphics[width=\textwidth]{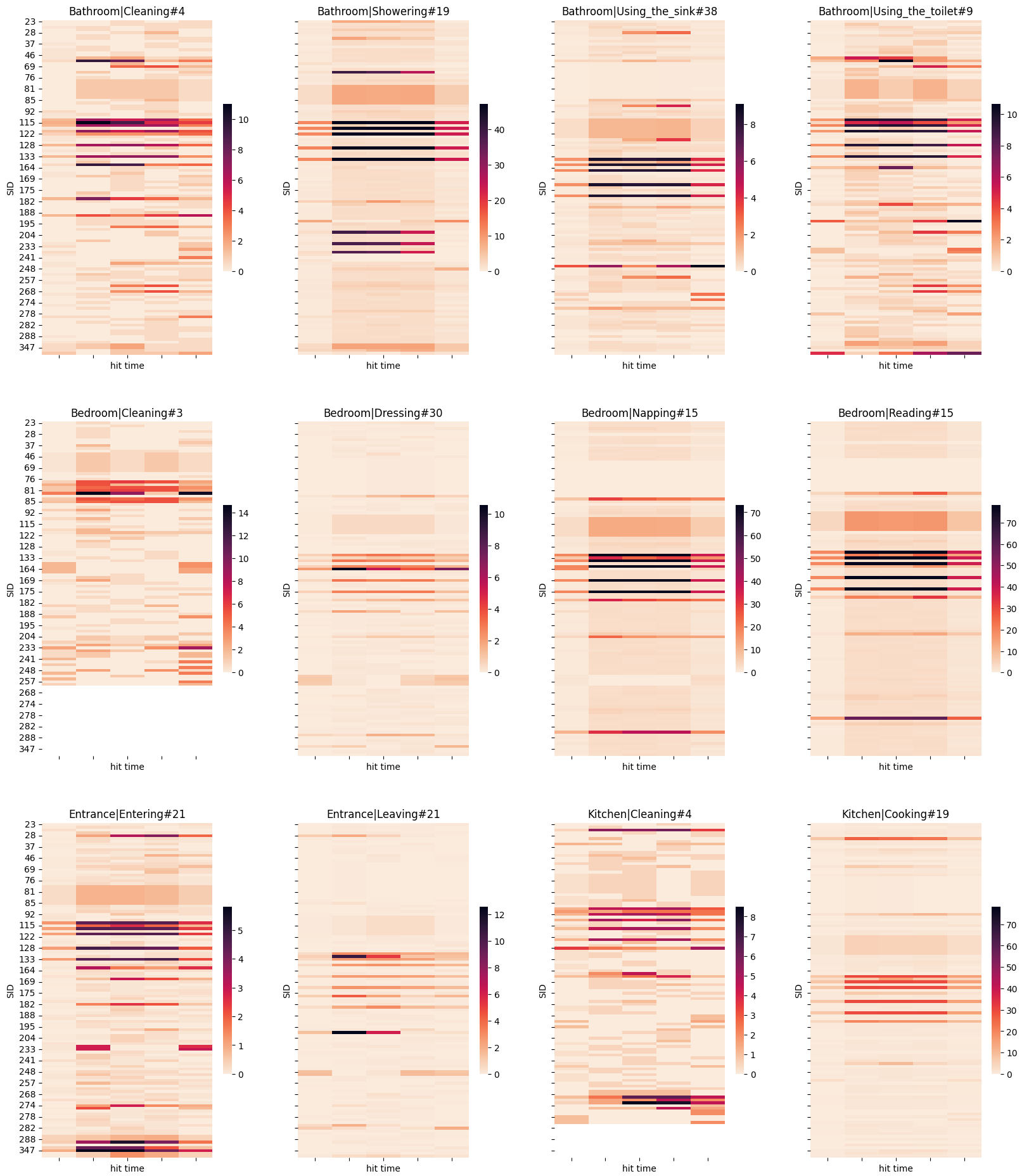}
    \caption[Sensor Event Frequency in Orange4Home Dataset (Part One)]{Sensor Event Frequency in the First Part (12 of 24 activities) of the Orange4Home Dataset. Each hit map indicates the frequency and distribution of sensor events for an individual activity, with the Y-axis representing specific sensors. The absence of events is shown in white, while darker tones indicate higher event frequencies. These visual patterns reveal significant differences in the occurrence and timing of sensor-triggered activities, such as between 'Kitchen Cooking' and 'Kitchen Cleaning'.
        Pre-segmented data showcases distinct patterns, reducing ambiguity and enhancing recognition performance of activities. Although valuable for experimental analysis, the practicality of pre-segmented data in real-world scenarios may be limited.}
    \label{fig:Orange4Home hit time1}
\end{figure}

\begin{figure}[tbh!]
    \centering
    \includegraphics[width=\textwidth]{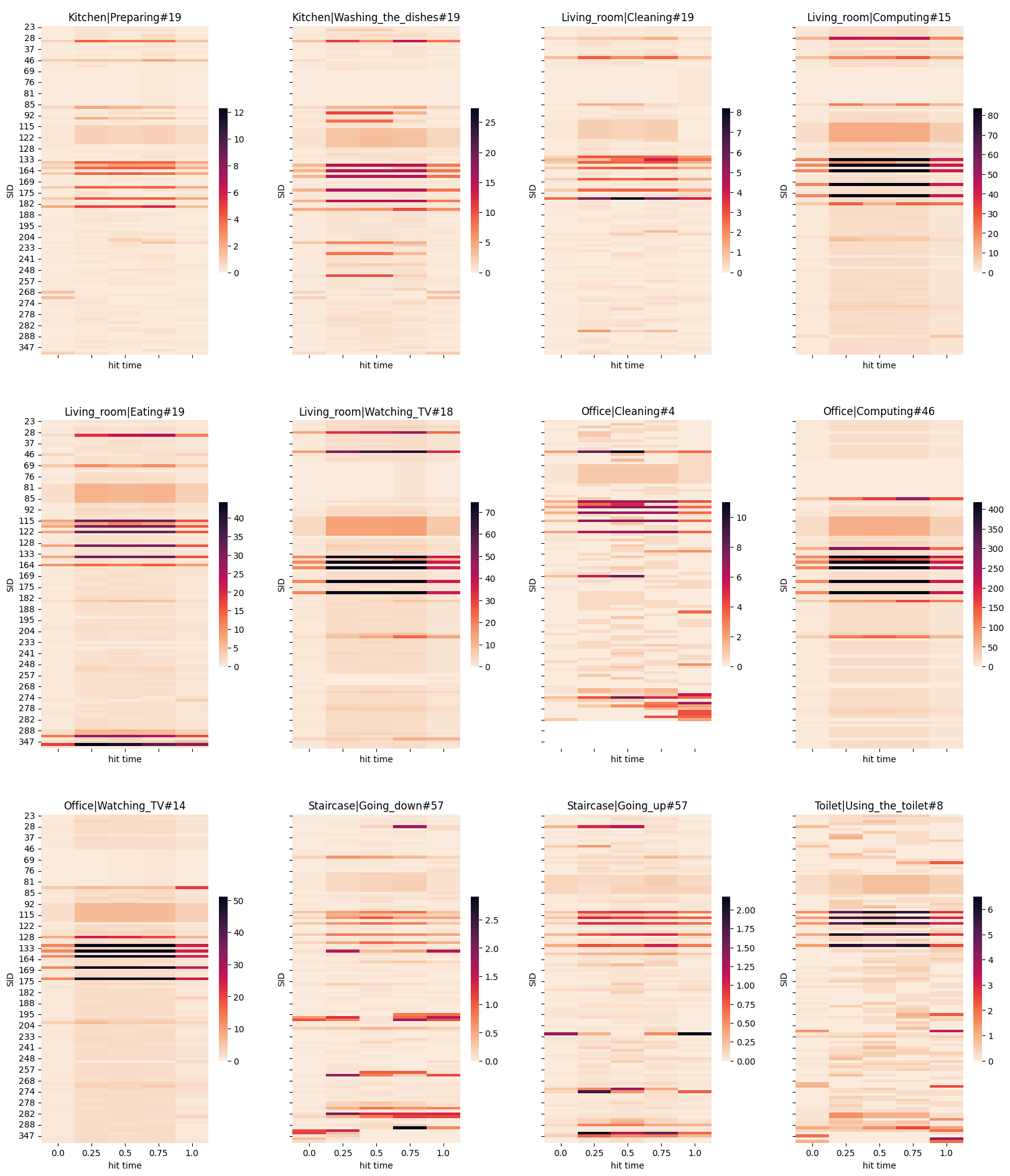}
    \caption[Sensor Event Frequency in Orange4Home Dataset (Part Two)]{Sensor Event Frequency for the remaining activities in the Orange4Home Dataset, complementing \cref{fig:Orange4Home hit time1}. Each hit map continues to represent the frequency and distribution of sensor events, with darker shades denoting more frequent occurrences. This part maintains the distinct activity patterns identified in the first half, supporting the more straightforward recognition of pre-segmented data across all 24 activities.}
    \label{fig:Orange4Home hit time2}
\end{figure}

\begin{figure}[h]
    \centering
    \includegraphics[width=.6\textwidth]{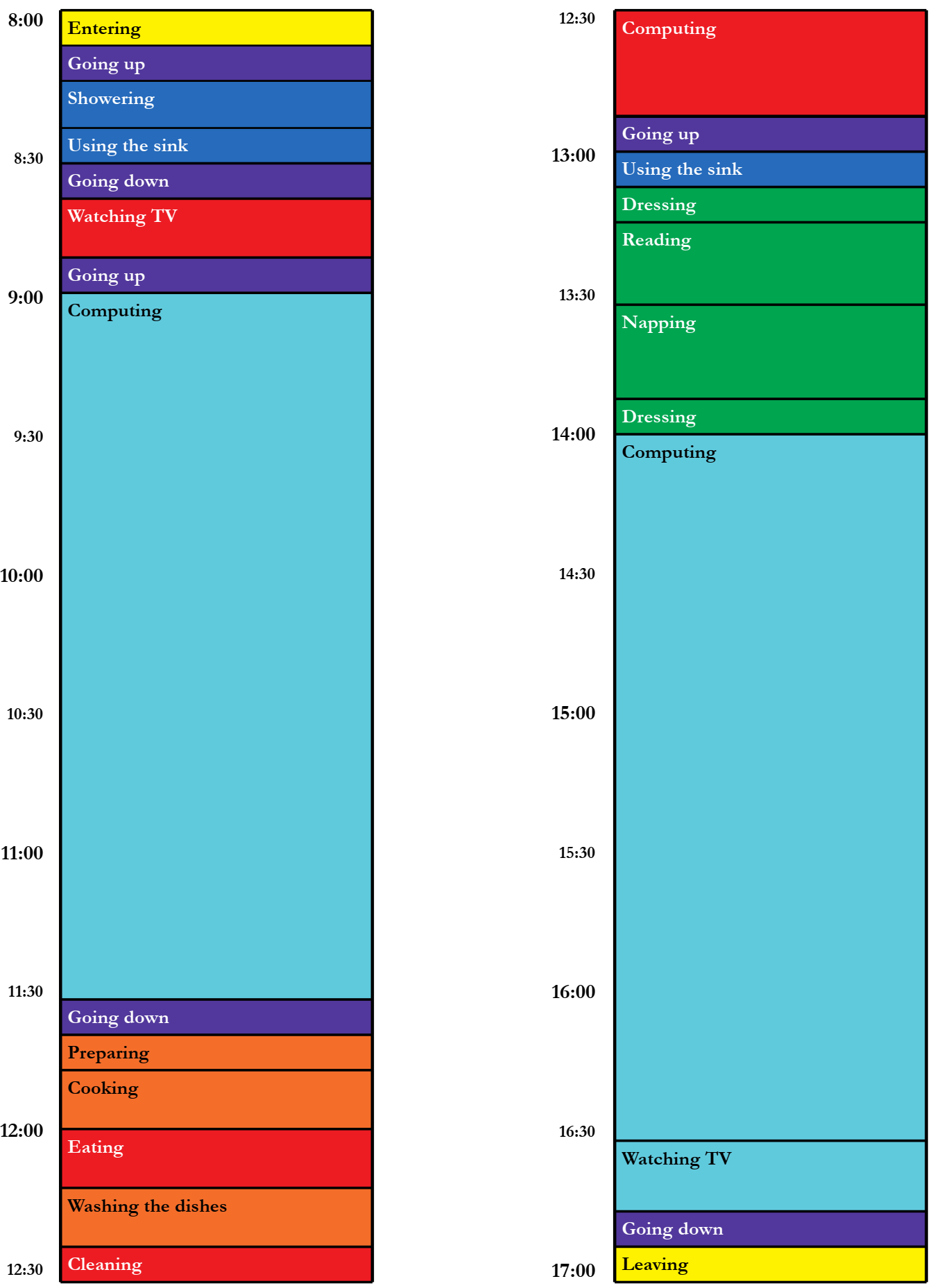}
    \caption{The standard routines in Orange4Home dataset \citep{Recognizing2018}}
    \label{fig:5:orange4home routines}
\end{figure}

\begin{figure}[t!]
    \centering
    \includegraphics[width=\textwidth]{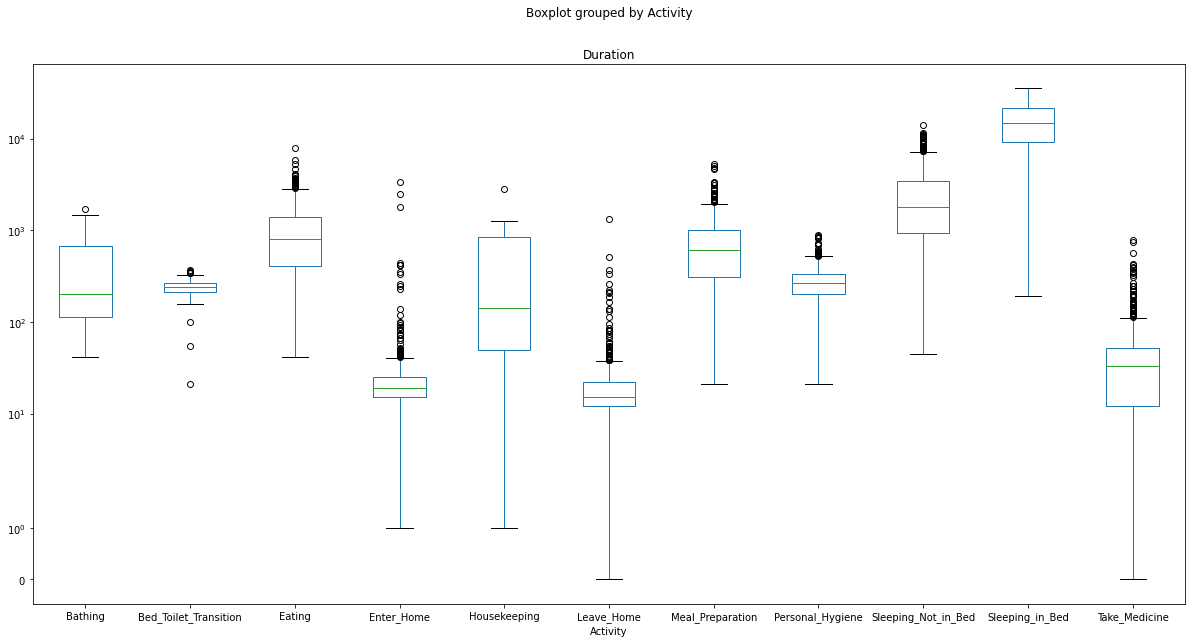}
    \caption{Home1 Activity Duration}
    \label{fig:home1 activity duration}
\end{figure}

\begin{figure}[t!]
    \centering
    \includegraphics[width=\textwidth]{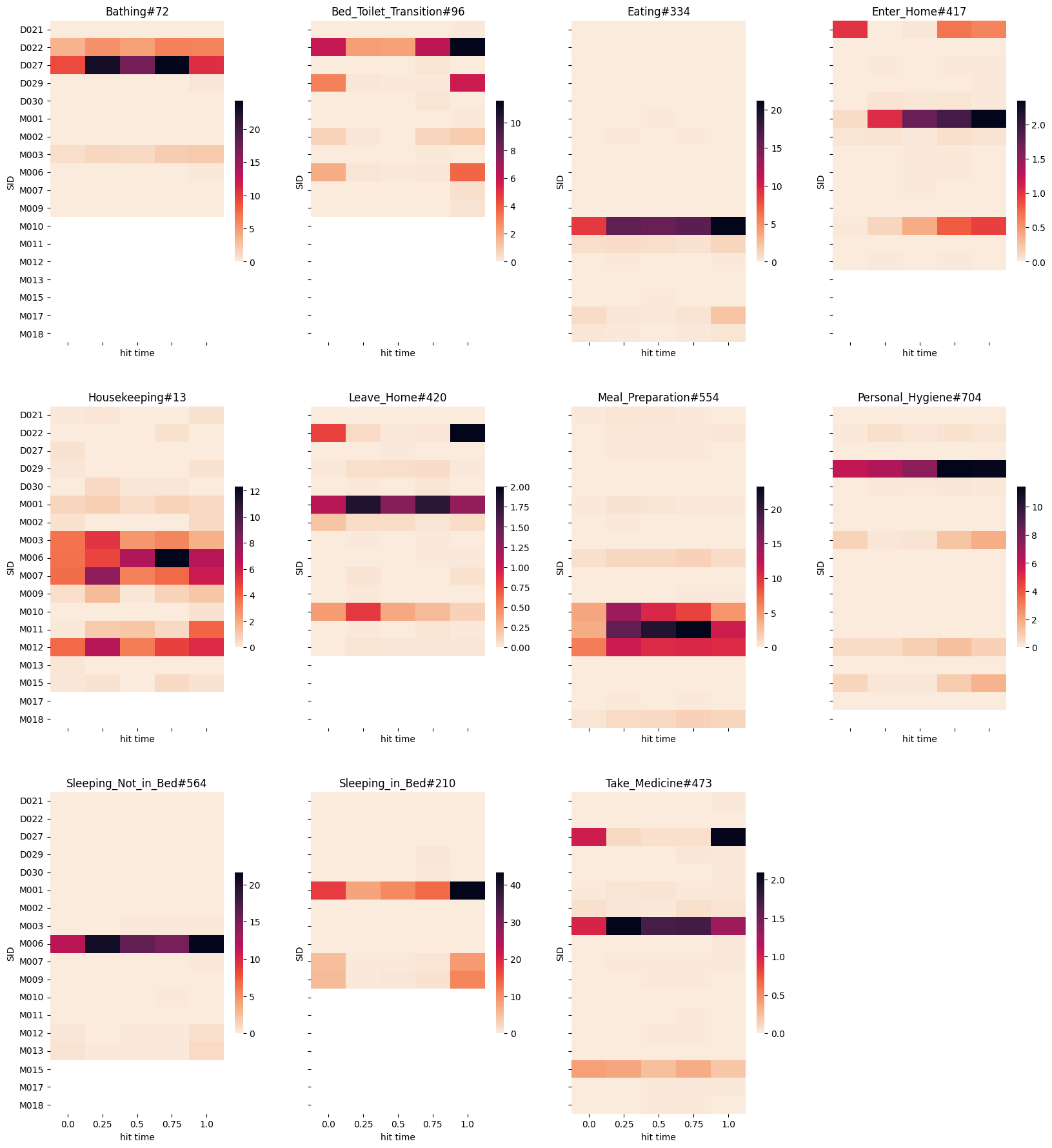}
    \caption{Sensor Event Frequency for activities in the Home1 Dataset. It shows the number of sensor events that occurred at each time for each activity. On the legend, the average number of sensor's heats are shown.}
    \label{fig:home1 hit time}
\end{figure}

\begin{figure}[t!]
    \centering
    \includegraphics[width=\textwidth]{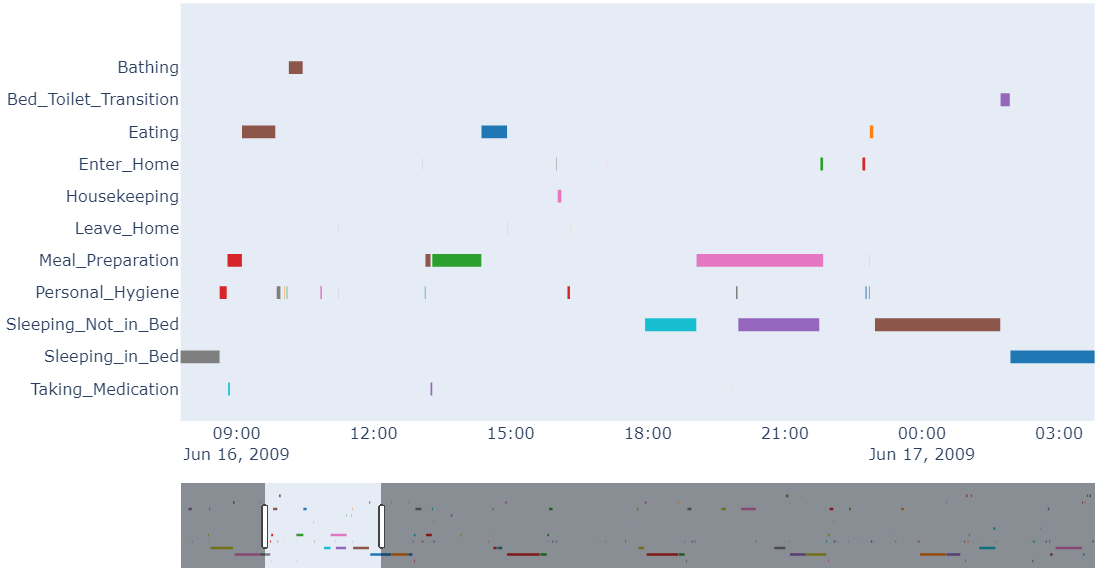}
    \caption{Home2 Activities}
    \label{fig:home2 activities}
\end{figure}

\begin{figure}[t!]
    \centering
    \includegraphics[width=\textwidth]{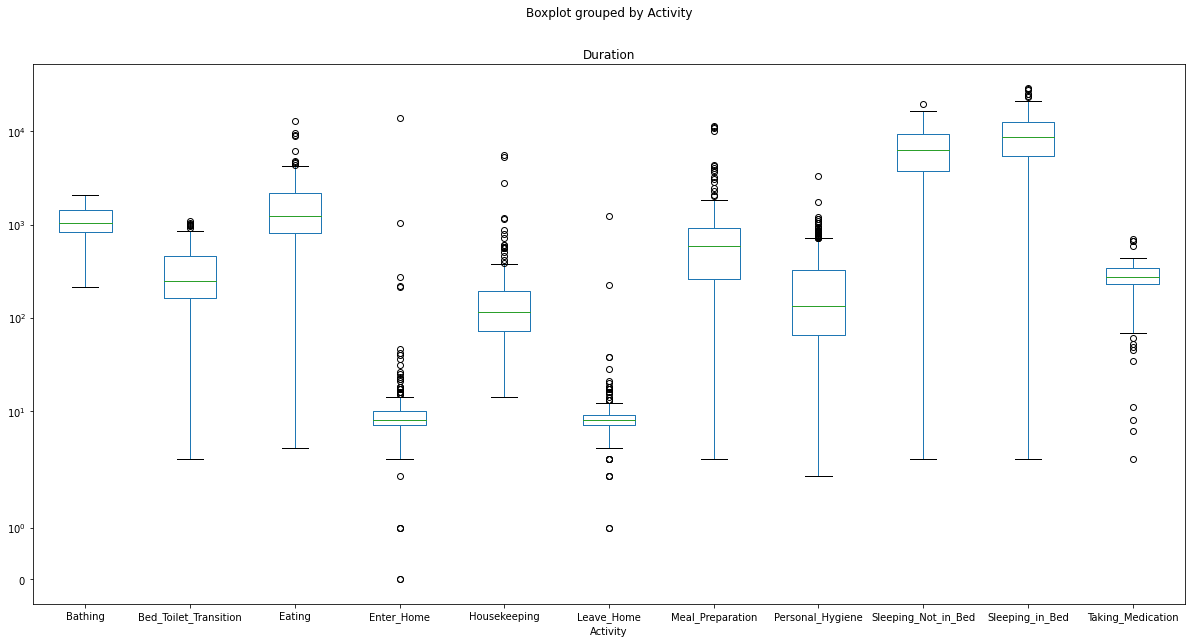}
    \caption{Home2 Activity Duration}
    \label{fig:home2 activity duration}
\end{figure}
\begin{figure}[t!]
    \centering
    \includegraphics[width=\textwidth]{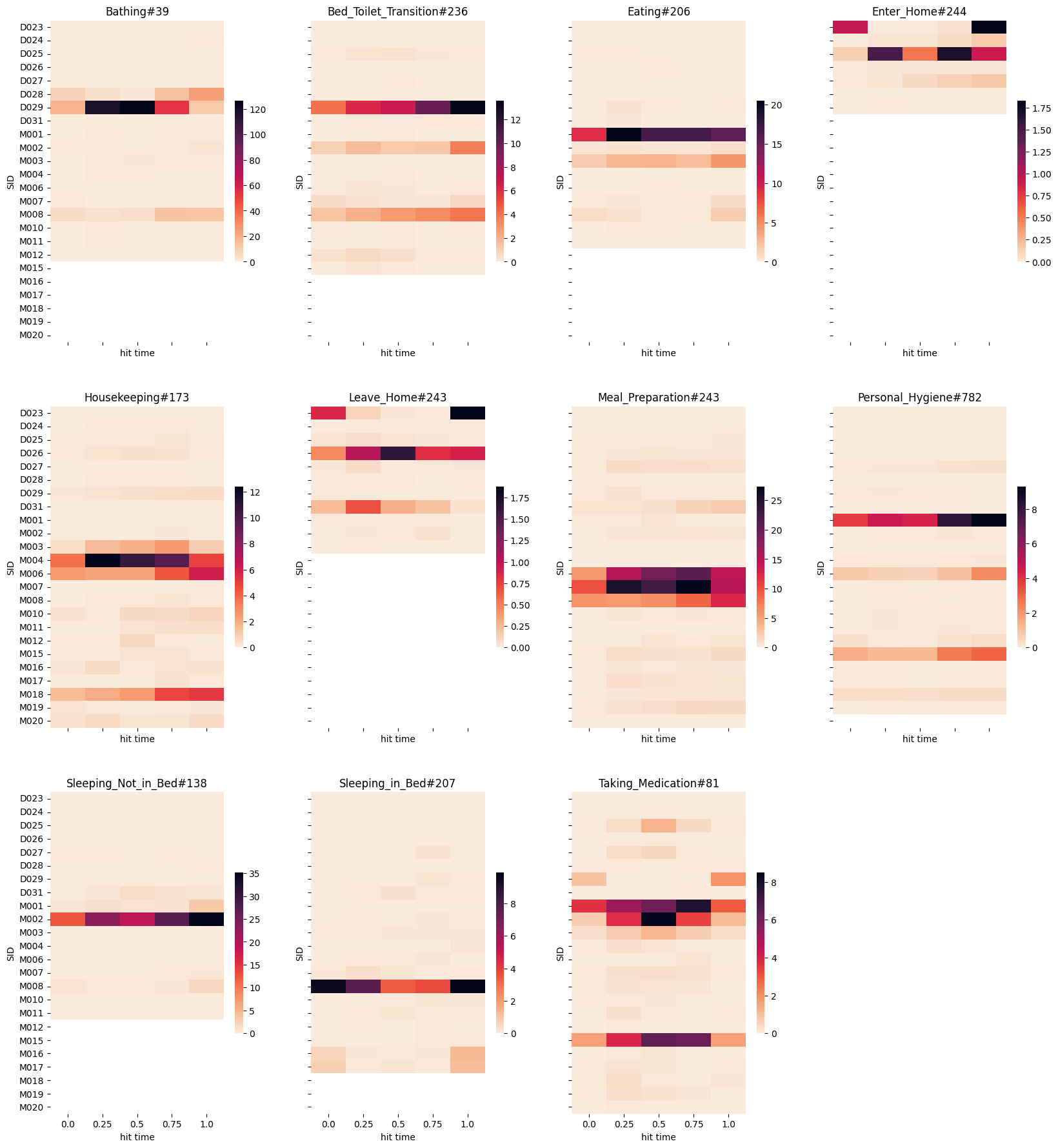}
    \caption{Sensor Event Frequency for activities in the Home2 Dataset. It shows the number of sensor events that occurred at each time for each activity. On the legend, the average number of sensor's heats are shown. }
    \label{fig:home2 hit time}
\end{figure}
\end{document}